\documentclass[11pt]{article}
\usepackage[margin=1in]{geometry}
\usepackage{booktabs,graphicx,amsmath,amssymb,hyperref,xcolor,microtype,natbib,listings,enumitem,tikz,tabularx,array}
\usetikzlibrary{arrows.meta,positioning,fit,calc,shapes.geometric}
\hypersetup{colorlinks=true,linkcolor=teal,citecolor=teal,urlcolor=teal}
\definecolor{codebg}{RGB}{238,242,242}
\definecolor{vblue}{HTML}{2A78D6}\definecolor{vorange}{HTML}{EB6834}\definecolor{vaqua}{HTML}{1BAF7A}
\definecolor{ink2}{HTML}{52514E}\definecolor{fillA}{HTML}{EEF4FB}\definecolor{fillB}{HTML}{FDF0EA}\definecolor{fillC}{HTML}{E8F6F0}
\newcommand{\site}{\url{https://menonpg.github.io/ivus-xa-registration/}}
\newcommand{\NRunsXA}{6}
\newcommand{\NFramesXA}{426}
\newcommand{\GreyMAE}{1.77}
\newcommand{\LongestAgree}{4}
\newcommand{\EngineFirst}{31--35}
\newcommand{\EngineWarm}{15}
\newcommand{\DecodeMFTwo}{5.3}
\newcommand{\DecodeMFEight}{6.8}
\newcommand{\DemoSec}{5.8}
\newcommand{\ExportSec}{1.1}
\newcommand{\ExportMB}{133}
\newcommand{\CenterlineSec}{3.33--5.96}
\newcommand{\AnalysisSec}{0.45--2.17}
\newcommand{\PhantomRuns}{36}
\newcommand{\TraceLumenMin}{87}
\newcommand{\TraceLumenMax}{99}
\newcommand{\TraceDistMin}{3.9}
\newcommand{\TraceDistMax}{4.6}
\newcommand{\LongestHi}{83}
\newcommand{\LongestHiNoisy}{54}
\newcommand{\RegShift}{17.5}
\newcommand{\SegLen}{90}
\newcommand{\PullbackFrames}{372}
\newcommand{\PullbackSec}{12.4}

\title{Every IVUS Frame in Its Place:\\[4pt]\large A Zero-Install Browser Tool for Registering Intravascular Ultrasound to the Coronary Angiogram, and a Path to Angiography-Only Plaque Imaging}
\author{Prahlad G. Menon\textsuperscript{1} \quad Srilakshmi Adhyapak\textsuperscript{2}\\[2pt]
\small \textsuperscript{1}The Menon Lab, University of Pittsburgh \quad \href{https://themenonlab.blog}{themenonlab.blog}\\
\small \textsuperscript{2}Department of Cardiology, St.\ John's Medical College Hospital, Bengaluru, India}
\date{November 2026}

\begin{document}
\maketitle

\begin{abstract}
\noindent\textbf{Background.}
Intravascular ultrasound (IVUS) shows what the coronary angiogram cannot: plaque composition, calcium arcs and stent apposition. Relating each IVUS cross-section to its position on the angiogram is still largely done in the operator's head. Commercial co-registration exists, but it is tied to vendor consoles and to set-up at acquisition time. Retrospective archives of exported angiograms and pullbacks therefore cannot be co-registered, and paired angiography--IVUS data for training artificial-intelligence (AI) models remain scarce.

\noindent\textbf{Methods.}
We built a semi-automatic registration tool that runs entirely in a web browser, with no installation and no upload. The clinician picks a contrast-filled reference frame, with help from an automatic contrast and cardiac-phase analysis. The tool proposes vessel centrelines from Frangi vesselness, which the clinician selects, edits or traces. The clinician then pins the first and last pullback frames, and any visible landmarks, to the centreline. A monotone piecewise-linear map then places every IVUS frame on the vessel. Scrubbing either view moves the other. Each case exports as an auditable training record. The original Python image analysis runs unchanged in WebAssembly, and videos are decoded frame-accurately in the browser. We measured fidelity against the original server pipeline on \NRunsXA{} angiographic runs, timed every operation, tested the automatic steps on \PhantomRuns{} phantom angiograms with known ground truth, and report a registered Left Main case.

\noindent\textbf{Results.}
The in-browser analysis was bit-identical to the server pipeline on identical frames. Browser decoding differed from ffmpeg by at most 2 grey levels (mean \GreyMAE{}). Suggested reference frames agreed in all \NRunsXA{} runs, and the default automatic centreline agreed in \LongestAgree{} of \NRunsXA{}. A 546-frame pullback decoded in \DecodeMFTwo{}\,s, and a complete case exported in \ExportSec{}\,s. On phantoms, the clinician-guided trace stayed inside the true lumen along \TraceLumenMin{}--\TraceLumenMax{}\% of the vessel in every noise and contrast condition. The fully automatic candidate fell from \LongestHi{}\% to 0\% as noise rose and contrast fell. In the clinical case, one side-branch landmark moved the position assigned to mid-pullback frames by \RegShift{}\% of the segment length compared with a constant-speed assumption.

\noindent\textbf{Conclusions.}
IVUS--angiogram registration can be done reproducibly by a clinician in an ordinary browser, on any archived case, without data leaving the computer. Automatic steps speed the work but must remain editable. Every registered case is a paired, frame-level training example. We describe how AI can enter at each step, from learned vessel segmentation to predicting a ``virtual IVUS'' from the angiogram alone. The tool is freely available online with a one-click demonstration case.
\end{abstract}

\begin{figure}[t]\centering
\includegraphics[width=\linewidth]{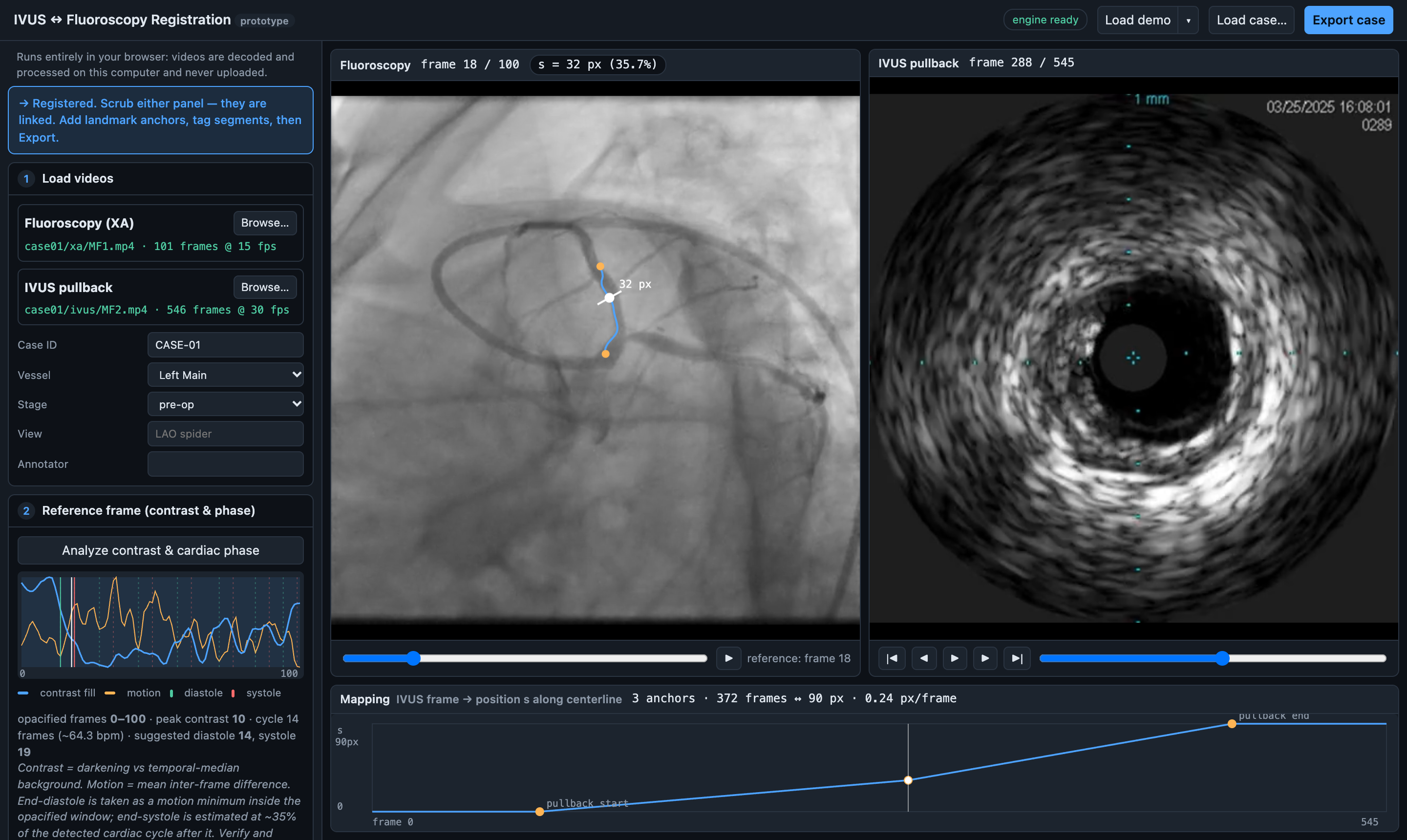}
\caption{The tool with the demonstration case loaded (Left Main, pre-intervention). Left: the five-step workflow. Centre: the contrast-filled reference angiogram, with the registered centreline (blue), the pinned frames (orange) and the position of the current IVUS frame (white marker, $s$ = 32\,px, 35.7\% along the segment). Right: the IVUS cross-section at that position (frame 288). Bottom: the mapping from IVUS frame to position along the vessel. Scrubbing the IVUS moves the marker; clicking the centreline moves the IVUS. Everything shown runs inside the browser. Try it at \site{} with \textbf{Load demo}.}
\label{fig:hero}
\end{figure}

\section{Introduction}
\label{sec:intro}
Coronary angiography shows the lumen as a projected silhouette. It shows little of what lies in the wall.
IVUS supplies the missing view: lumen and vessel areas, plaque burden, calcium arcs, and the expansion and apposition of a stent \citep{mintz2001acc,raber2018eapci}.
IVUS guidance improves outcomes after stenting compared with angiography alone \citep{zhang2018ultimate,lee2023renovate}.
Angiography also underestimates calcium. In a classic series of 1{,}155 lesions, IVUS detected calcium far more often than the angiogram \citep{mintz1995calcification}.

The two modalities are acquired at different times and in different geometries.
To use them together, the operator has to know where each IVUS cross-section lies on the angiogram.
In routine practice this is done by eye, from the pullback speed and from landmarks such as side branches.
Commercial co-registration systems automate it, but only when the IVUS console, the X-ray system and the acquisition protocol are set up for it.
For the large archives of angiograms and pullbacks that are exported as ordinary video files, no such link exists.

This gap matters twice. Clinically, a reviewer or trainee cannot easily answer ``what does the vessel look like \emph{here}?'' on an archived case.
For research, it blocks a promising question.
Angiography-derived fractional flow reserve showed that physiology can be computed from the angiogram alone \citep{tu2016favor}.
Can plaque morphology, the content of an IVUS frame, also be predicted from the angiogram?
Answering it requires many angiograms in which every IVUS frame is tied to a position on the vessel, and such paired data barely exist.

We present a tool that creates exactly this pairing, designed with an interventional cardiologist (S.A.) for clinicians who are not programmers. Our contributions are:
\begin{enumerate}[nosep]
\item a semi-automatic, clinician-in-the-loop registration workflow with linked, bidirectional angiogram--IVUS viewing (Figure~\ref{fig:hero});
\item a browser edition that needs no installation and uploads nothing. The original Python image analysis runs unchanged in WebAssembly, and videos are decoded frame-accurately by the browser;
\item a measured evaluation: numerical fidelity against the server pipeline, timing, accuracy of the automatic steps on phantoms with known ground truth, and a registered clinical case;
\item an export format that turns each registered case into a frame-level angiography--IVUS training record, and a one-click demonstration case;
\item a map of where AI can enter the workflow, including the end goal of angiography-only ``virtual IVUS''.
\end{enumerate}

\section{Background and related work}
\label{sec:related}

\paragraph{Co-registration and fusion.}
Fusing IVUS with angiography has a long history in research software.
Biplane angiography can position each IVUS frame in three dimensions \citep{wahle1999fusion}, and 3D quantitative coronary angiography has been fused with IVUS and OCT \citep{tu2011fusion}.
These methods aim at geometric accuracy and need calibrated biplane or multi-view acquisitions.
Our goal is different and complementary: a lightweight, single-view, clinician-driven registration that works on whatever an archive contains and produces training data.

\paragraph{Vessel enhancement and segmentation.}
Classical pipelines enhance tubular structures with multiscale Hessian ``vesselness'' \citep{frangi1998vesselness}, then threshold and skeletonise the result \citep{zhang1984thinning}.
Minimal-path methods trace a vessel between two points \citep{dijkstra1959,cohen1997minimal}.
Deep networks now segment the major coronary arteries \citep{ronneberger2015unet,yang2019angiosegmentation}, and segment IVUS borders in real time \citep{bajaj2021ivusseg}.
We deliberately ship the classical pipeline and keep it swappable (\S\ref{sec:ai}), because it needs no training data and its failures are easy to see and correct.

\paragraph{Annotation tools and running code in the browser.}
General platforms such as 3D Slicer \citep{fedorov2012slicer} can annotate both modalities but need installation, configuration and data handling that a busy clinic rarely has.
Two recent browser technologies remove those barriers.
Pyodide compiles CPython and the scientific Python stack (NumPy, SciPy, scikit-image, OpenCV; \citealp{harris2020numpy,virtanen2020scipy,vanderwalt2014skimage,bradski2000opencv}) to WebAssembly \citep{pyodide}.
WebCodecs gives web pages frame-accurate access to the hardware video decoder \citep{webcodecs}.
Together they let a validated Python pipeline run where the data already are, in the clinician's browser.

\section{Methods}
\label{sec:methods}

\subsection{Requirements}
The requirements came from joint review sessions on a real Left Main case. In those sessions, deciding even the direction of the pullback took several minutes of discussion.
\begin{enumerate}[nosep,label=R\arabic*]
\item \textbf{Frame accuracy.} Every IVUS frame and every angiogram frame must be addressable by index, exactly as stored.
\item \textbf{Clinician in control.} Automatic steps may propose; every result must be editable, and drawing by hand must always be possible.
\item \textbf{Linked viewing.} Moving through the IVUS moves a marker along the vessel, and clicking the vessel moves the IVUS.
\item \textbf{Uneven pullback.} Landmarks pinned by the clinician must bend the mapping, which must stay monotone because the probe does not move backwards.
\item \textbf{Reproducible output.} An exported case must reconstruct the registration without the tool.
\item \textbf{No data movement, no installation.} The tool must run on a standard hospital or laptop browser, and images must not leave the machine.
\end{enumerate}

\subsection{Clinical workflow}
\label{sec:workflow}
The interface (Figure~\ref{fig:hero}) guides the user through five numbered steps. A banner always states the next action, and a locked step says why it is locked.

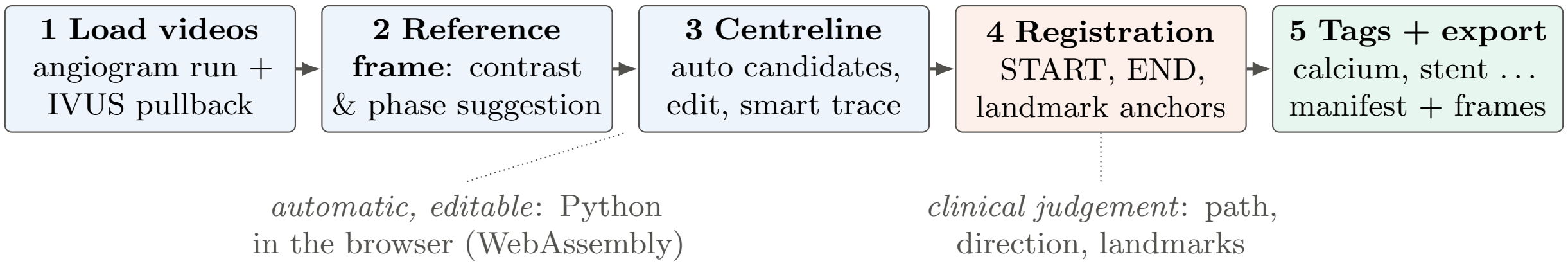
\begin{figure}[t]\centering
\begin{tikzpicture}[node distance=3.5mm and 2.6mm, font=\footnotesize,
  step/.style={draw=ink2, rounded corners=2pt, fill=#1, align=center, minimum height=13mm, text width=28.5mm, inner sep=2pt},
  arr/.style={-{Latex[length=2mm]}, thick, ink2}]
\node[step=fillA] (s1) {\textbf{1 Load videos}\\angiogram run +\\IVUS pullback};
\node[step=fillA, right=of s1] (s2) {\textbf{2 Reference}\\\textbf{frame}: contrast\\\& phase suggestion};
\node[step=fillA, right=of s2] (s3) {\textbf{3 Centreline}\\auto candidates,\\edit, smart trace};
\node[step=fillB, right=of s3] (s4) {\textbf{4 Registration}\\START, END,\\landmark anchors};
\node[step=fillC, right=of s4] (s5) {\textbf{5 Tags + export}\\calcium, stent \ldots\\manifest + frames};
\foreach \a/\b in {s1/s2,s2/s3,s3/s4,s4/s5} \draw[arr] (\a) -- (\b);
\node[below=5mm of s2, text width=62mm, align=center, text=ink2] (auto) {\emph{automatic, editable}: Python in the browser (WebAssembly)};
\node[below=5mm of s4, text width=46mm, align=center, text=ink2] (clin) {\emph{clinical judgement}: path, direction, landmarks};
\draw[ink2, densely dotted] (auto.north) -- ($(s2.south)!.5!(s3.south)$);
\draw[ink2, densely dotted] (clin.north) -- (s4.south);
\end{tikzpicture}
\caption{The five-step workflow. Blue steps are automated but always editable; orange is where the clinician's knowledge is essential; green produces the training record.}
\label{fig:workflow}
\end{figure}

\begin{figure}[t]\centering
\includegraphics[width=.62\linewidth]{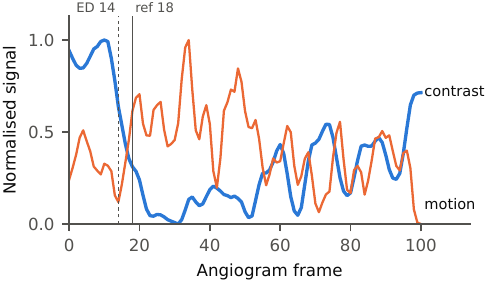}
\caption{Contrast filling (blue) and frame-to-frame motion (orange) for angiographic run MF1. The tool suggested frame 14 as end-diastole (dotted); the reference used for registration was frame 18 (solid), a few frames later, chosen by the clinician for vessel filling.}
\label{fig:phase}
\end{figure}

\begin{enumerate}[nosep]
\item \textbf{Load videos.} The user opens a folder of exported runs. Each run is listed with a thumbnail, frame count, frame rate and a label (``pullback / long run'', ``short run'', ``still''), so the pullback is easy to identify (Appendix Figure~\ref{fig:browse}). Case metadata are entered here: vessel, pre- or post-intervention stage, view and annotator.
\item \textbf{Reference frame.} An analysis of the angiographic run plots contrast filling and frame-to-frame motion, and suggests an end-diastolic and an end-systolic frame (Figure~\ref{fig:phase}). The user inspects the suggestion and freezes the chosen frame.
\item \textbf{Centreline.} The tool proposes candidate centrelines (Figure~\ref{fig:pipeline}c). The user picks the one that follows the pullback path, or uses \emph{smart trace}: two clicks, and a minimal-cost path follows the vessel between them. Control points can be dragged, inserted or deleted. The line is trimmed to the pulled-back segment and, if needed, flipped so that position $s=0$ is where the pullback started.
\item \textbf{Registration.} The user pins the first and last pullback frames to the two ends of the centreline. Landmarks visible in both images, such as a side-branch ostium or a stent edge, are added as anchors. The new anchor's point on the mapping chart is dragged until its dot on the angiogram sits on the landmark.
\item \textbf{Tags and export.} IVUS frame ranges are tagged (calcium, fibrous, lipid, stent, bifurcation, partial or circumferential extent), and the case is exported.
\end{enumerate}

\subsection{Algorithms}
\label{sec:algorithms}
\paragraph{Contrast and cardiac phase.}
Frames are downscaled to at most 256 pixels per side.
Because contrast makes vessels darker, each pixel's 85th temporal percentile serves as an un-opacified background $B$.
The contrast signal of frame $t$ is the fraction of pixels darker than $B$ by more than 10 grey levels, multiplied by their mean darkening.
The motion signal is the mean absolute difference between consecutive frames.
Both are smoothed and normalised to $[0,1]$.
The opacified window is the range of frames with contrast $\geq 0.45$.
The cardiac period is the autocorrelation peak of the motion signal between 0.4 and 1.6\,s.
End-diastolic candidates are motion minima inside the window, ranked by $\text{contrast}\times(1-\text{motion})$.
End-systole is placed 35\% of a cycle later, because no ECG is available in exported videos.
These are explicit heuristics, displayed with their rationale and always overridable.

\paragraph{Centreline candidates.}
On the reference frame, the background-subtracted image $\max(B-I,0)$ is blurred ($\sigma=1$) and normalised. Frangi vesselness \citep{frangi1998vesselness} is then computed at five scales between 1 and 7 pixels.
The vesselness map is thresholded at its $(99-6\,\kappa)$th percentile, where $\kappa\in[0,1]$ is the user's sensitivity slider. It is then closed morphologically, and components smaller than 12\% of the largest are removed. Finally it is skeletonised \citep{zhang1984thinning}, and spurs shorter than about 12 pixels are pruned.
The skeleton is split into segments between junctions and end points.
The default ``longest path'' candidate is the diameter of a maximum-weight spanning forest of the segment graph \citep{kruskal1956}, found by a double graph sweep. This avoids the intractable longest-path problem on graphs with cycles.
It is offered together with the 12 longest individual segments. Every candidate is smoothed and resampled every 2 pixels.

\paragraph{Smart trace.}
Between two clicked points, the tool finds the minimal-cost 8-connected path \citep{dijkstra1959,cohen1997minimal} over the cost $1/(v+0.02)$, where $v$ is the vesselness.

\paragraph{Registration map.}
Let the centreline have arc length $L$, and let the anchors be pairs $(f_k, u_k)$ of IVUS frame and fractional position $u_k=s_k/L$, sorted by frame.
Positions are made monotone by a running maximum, $\tilde u_k=\max_{j\le k}u_j$. Every frame $f$ is then mapped by linear interpolation, and held constant outside the first and last anchors:
\begin{equation}
u(f)=\tilde u_k+\frac{f-f_k}{f_{k+1}-f_k}\,(\tilde u_{k+1}-\tilde u_k),\qquad f_k\le f\le f_{k+1}.
\label{eq:map}
\end{equation}
With only START and END this is the familiar constant-speed assumption. Each further landmark corrects the speed on either side of it.
The inverse map, from a clicked position to a frame, drives the linked view.
Storing fractions rather than pixels keeps anchors valid if the centreline is re-smoothed. An optional calibration (pixels per mm) converts positions to millimetres.

\subsection{A browser edition with no installation and no upload}
\label{sec:arch}
\begin{figure}[t]\centering
\begin{tikzpicture}[font=\footnotesize, box/.style={draw=ink2, rounded corners=2pt, fill=#1, align=left, inner sep=4pt, text width=60mm, minimum height=24mm},
  arr/.style={-{Latex[length=2mm]}, thick, ink2}]
\node[box=fillA, anchor=north west] (ui) at (0,0) {\textbf{Page (main thread)}\\[1pt]
interface, linked viewers, mapping chart\\
\textbf{WebCodecs} + mp4box.js: every frame\\
frames kept as lossless PNG in memory\\
JSZip: \texttt{case.zip} to Downloads};
\node[box=fillC, anchor=north west] (wk) at (84mm,0) {\textbf{Web worker}\\[1pt]
\textbf{Pyodide}: CPython in WebAssembly\\
NumPy, SciPy, scikit-image, OpenCV\\
\texttt{analysis.py}, \texttt{centerline.py}\\
unchanged from the server version};
\node[draw=ink2, rounded corners=2pt, fill=fillB, align=center, inner sep=4pt, text width=140mm, anchor=south] (disk) at ($(ui.north west)!.5!(wk.north east)+(0,9mm)$)
  {\textbf{Clinician's computer}: video files opened with a file picker; nothing is uploaded};
\node[draw=ink2, dashed, rounded corners=2pt, align=center, inner sep=4pt, text width=140mm, anchor=north] (net) at ($(ui.south west)!.5!(wk.south east)+(0,-9mm)$)
  {\textbf{From the web}: static site (GitHub Pages) and libraries from public CDNs: program code only, never images};
\draw[arr] (ui.north |- disk.south) -- (ui.north);
\draw[arr] ([yshift=4mm]ui.east) -- node[above, font=\scriptsize]{grey frames} ([yshift=4mm]wk.west);
\draw[arr] ([yshift=-4mm]wk.west) -- node[below, font=\scriptsize]{results} ([yshift=-4mm]ui.east);
\draw[arr, dashed] (net.north -| ui.south) -- (ui.south);
\draw[arr, dashed] (net.north -| wk.south) -- (wk.south);
\end{tikzpicture}
\caption{Browser architecture. Image data flow only between the user's disk, the page and a background worker on the same computer. The network supplies program code only.}
\label{fig:arch}
\end{figure}
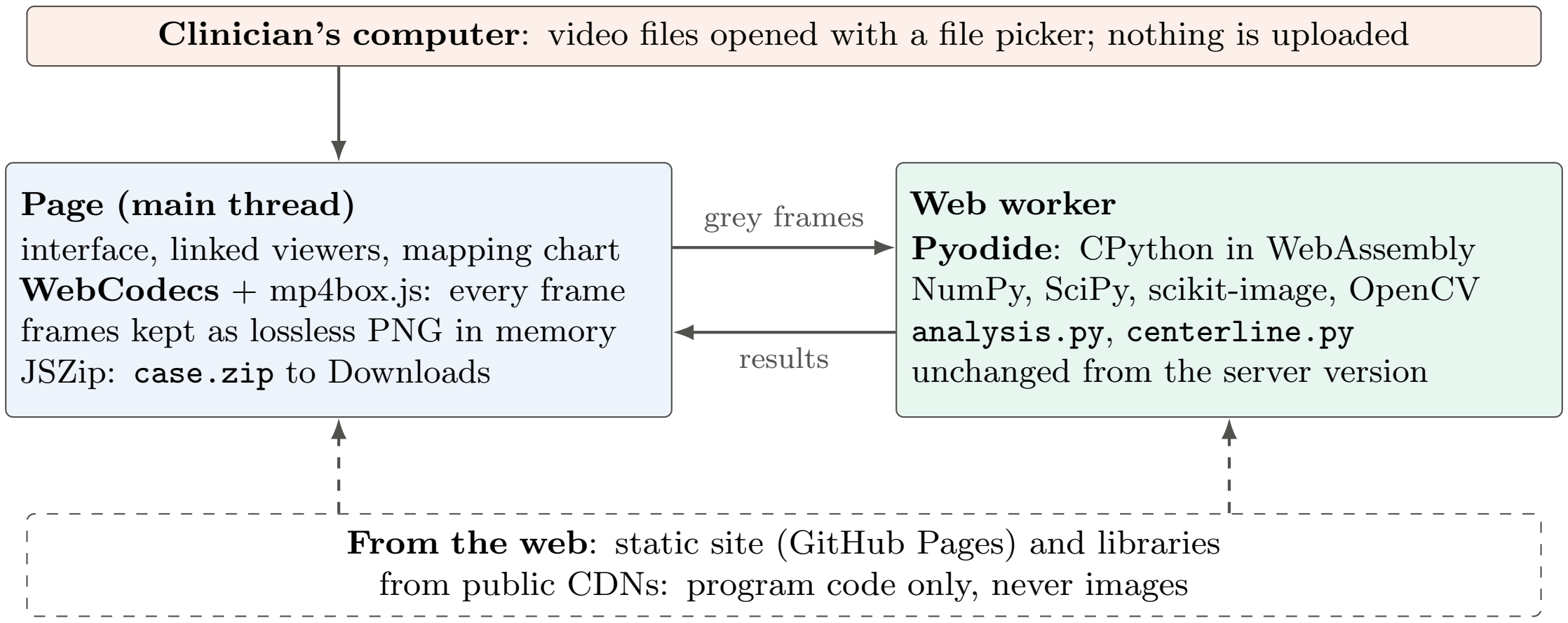

The tool was first written as a local server (FastAPI) that extracted frames with ffmpeg and analysed them in Python.
For deployment we ported it to the browser without rewriting the image analysis (Figure~\ref{fig:arch}).

\paragraph{Frame-accurate decoding.}
MP4 and MOV files are demultiplexed with mp4box.js \citep{mp4boxjs}, and every coded sample is decoded with WebCodecs \citep{webcodecs}. Output is ordered by presentation timestamp.
This yields exactly one image per stored frame, the behaviour of \texttt{ffmpeg -vsync 0} used by the server version.
Frames are stored as lossless PNG for display and export.
For angiograms, a grey-level stack is also computed with OpenCV's fixed-point RGB-to-grey coefficients, so it matches the server's \texttt{cv2.imread} values except for decoder differences (\S\ref{sec:res-fidelity}).

\paragraph{Python in the browser.}
A web worker loads Pyodide 0.27.7 and the NumPy, SciPy, scikit-image and OpenCV packages.
It then runs the server's \texttt{analysis.py} and \texttt{centerline.py} with one change: frames are read from memory instead of from disk.
The page talks to the worker through the same function names the server exposed as HTTP endpoints, so the interface code barely changed.

\paragraph{Privacy by construction.}
Videos are opened with the browser's file picker and never transmitted. The site is static (GitHub Pages) and has no server-side code that could receive data.
The only network traffic is the program and its libraries.
This removes the data-transfer question that usually blocks web tools in hospitals, though local policy on software use still applies.

\subsection{Export format and demonstration mode}
Each case exports as a zip containing the reference angiogram frame, every IVUS frame (lossless PNG) and a JSON manifest (Appendix~\ref{app:manifest}).
The manifest holds the case metadata, the centreline points with their cumulative arc length, the anchors with both pixel and fractional positions, the per-frame map $f\mapsto s$ for every IVUS frame, and the tags with their positions along the vessel.
This is everything a model needs to learn ``angiogram location $\to$ IVUS appearance'', and it reconstructs the registration without the tool (R5).
A \emph{demo mode} loads bundled videos and a saved registration in one click (\texttt{?demo} in the address), for teaching and for reviewers. A demo file format (Appendix~\ref{app:demo}) lets any group package its own case.

\subsection{Data}
\label{sec:data}
\paragraph{Clinical case.}
One de-identified Left Main case (Case~01) was available: six angiographic runs (50--101 frames, 15 frames/s) and nine IVUS series (1--895 frames, 30 frames/s), all 500$\times$500 pixels and H.264-encoded.
Files were renamed and container metadata removed, with no re-encoding; decoded frames were verified to be bit-identical to the originals.
The images contain no patient identifiers.

\paragraph{Phantoms.}
To measure accuracy against ground truth, we generated angiogram phantoms (512$\times$512 pixels, 75 frames, 15 frames/s; Figure~\ref{fig:phantom}a). Each contains a main vessel (radius 8 tapering to 4.5 pixels) with two side branches, a guiding catheter, a textured background with a cardiac shadow, contrast wash-in from frame 6 to 18 with wash-out after frame 50, and periodic cardiac motion (period 13 frames, 3.5\% scaling plus a 3--4 pixel shift).
Contrast was set to 85, 40 or 20 grey levels of darkening, and Gaussian noise to a standard deviation of 2, 4, 8 or 16 grey levels, with three random seeds each (36 phantoms).
A matching synthetic IVUS pullback (360 frames) with a focal eccentric stenosis, a calcified arc with acoustic shadow and a side-branch take-off is used for the public demonstration (Figure~\ref{fig:phantom}b, Appendix Figure~\ref{fig:synthetic}).

\subsection{Evaluation}
\begin{description}[nosep,leftmargin=0pt]
\item[E1, port fidelity.] On all \NRunsXA{} angiographic runs (\NFramesXA{} frames), the server pipeline and the in-browser Python were run on the same ffmpeg-decoded frames. Every output was compared byte for byte.
\item[E2, browser end to end.] The complete browser pipeline, with WebCodecs decoding, was run in Chrome~157 on a 2021 MacBook Pro (Apple M1 Max, 64\,GB). Its grey levels, suggested frames and centreline candidates were compared with the server's, and every operation was timed.
\item[E3, automatic steps on phantoms.] For each phantom we scored: the detected opacified window against the true one (frames 12--66); the true opacification of the suggested reference frame; the smart trace between the true vessel ends; and the default longest-path candidate and best single candidate. Trace and candidates were scored as the share of the true centreline whose nearest traced point lies inside the local lumen radius. As an ablation, we repeated the experiment with background subtraction disabled.
\item[E4, case study.] We report the clinician's registration of Case~01 and the effect of its interior landmark.
\end{description}

\section{Results}
\label{sec:results}

\subsection{The port is exact; the decoders differ by two grey levels}
\label{sec:res-fidelity}
On identical frames, the in-browser Python reproduced the server pipeline bit for bit in all \NRunsXA{} runs (Table~\ref{tab:fidelity}): contrast and motion curves, suggested frames, all 13 centreline candidates per run, smart traces and vesselness images.
WebCodecs and ffmpeg convert colour slightly differently: every decoded pixel differed by 1 or 2 grey levels (mean \GreyMAE{}), and none was identical.
Despite this, the suggested end-diastolic and end-systolic frames, the opacified window and the cardiac period agreed in every run.
The default longest-path candidate agreed within 1.2 pixels of length and 1 pixel at its ends in \LongestAgree{} runs. In the other two it ended on a different branch (MF6) or took a different route (MF2).
The automatic candidates are therefore sensitive to small image changes, a further reason why the clinician chooses among them (R2).

\begin{table}[t]\centering\small
\caption{Fidelity of the browser edition on the six angiographic runs of Case~01. \emph{Port} = the in-browser Python on the server's decoded frames: contrast and phase curves, all centreline candidates, the smart trace and the vesselness image were bit-identical to the server pipeline (\checkmark). The remaining columns compare the complete browser run (WebCodecs decoding) with the server run: mean absolute grey-level difference of the decoded frames (maximum 2 levels in every run), whether the suggested end-diastolic and end-systolic frames agree, the number of automatic candidates, the length of the default ``longest path'' candidate (browser / server, pixels) and the largest shift of its end points.}
\label{tab:fidelity}
\begin{tabular}{lrcrcccr}\toprule
Run & Frames & Port & Grey MAE & Same frames & Candidates & Longest path (px) & End shift (px) \\ \midrule
MF1 & 101 & \checkmark & 1.76 & yes & 13 / 13 & 275.6 / 275.1 & 0 \\
MF2 & 101 & \checkmark & 1.77 & yes & 13 / 13 & 597.6 / 675.8 & 0 \\
MF3 & 62 & \checkmark & 1.77 & yes & 13 / 13 & 192.3 / 192.3 & 0 \\
MF4 & 62 & \checkmark & 1.77 & yes & 13 / 13 & 191.1 / 191.2 & 0 \\
MF5 & 50 & \checkmark & 1.77 & yes & 13 / 13 & 406.9 / 405.7 & 1 \\
MF6 & 50 & \checkmark & 1.76 & yes & 13 / 13 & 417.2 / 416.9 & 128 \\
\bottomrule\end{tabular}
\end{table}

\subsection{Speed}
\label{sec:res-speed}
Table~\ref{tab:runtime} lists timings.
On a first visit the Python engine took \EngineFirst{}\,s to download and start, and \EngineWarm{}\,s on later visits. It loads in the background while the user opens videos.
A 546-frame pullback was decoded and stored in \DecodeMFTwo{}\,s, and an 895-frame pullback in \DecodeMFEight{}\,s.
Phase analysis took \AnalysisSec{}\,s per run and centreline candidates \CenterlineSec{}\,s. That is 3--5$\times$ and 8--14$\times$ slower than native Python on the same machine, but still only a few seconds for a step done once per case. A smart trace took under 0.2\,s.
The one-click demonstration loaded both videos and the saved registration in \DemoSec{}\,s, and a complete case exported in \ExportSec{}\,s (\ExportMB{}\,MB).
Decoding slowed several-fold while the browser tab was in the background (an angiographic run of 50 frames: 0.65\,s in the foreground, 2.4\,s in the background), because browsers deprioritise hidden pages. The analysis and centreline timings were measured with the tab in the background and are therefore upper bounds.

\begin{table}[t]\centering\small
\caption{Time per operation in Chrome on a 2021 MacBook Pro (Apple M1 Max, 64\,GB), Case~01 data. Ranges are over the six angiographic runs (50--101 frames, 500$\times$500 pixels). The native-Python column runs the same code outside the browser on the same machine, for reference. IVUS decoding was timed with the tab in the foreground; analysis and centreline timings were taken with the tab in the background and are upper bounds.}
\label{tab:runtime}
\begin{tabular}{lrr}\toprule
Operation & Browser & Native Python \\ \midrule
Image engine ready, first visit (download + start) & 31--35\,s & -- \\
Image engine ready, later visits (cached) & 15\,s & -- \\
Decode IVUS pullback, 546 frames (5.3\,MB MP4) & 5.3\,s & -- \\
Decode IVUS pullback, 895 frames (8.8\,MB MP4) & 6.8\,s & -- \\
Contrast and cardiac-phase analysis & 0.45--2.17\,s & 0.16--0.40\,s \\
Automatic centreline candidates & 3.33--5.96\,s & 0.42--0.46\,s \\
Smart trace between two clicks & 0.04--0.16\,s & -- \\
One-click demo (both videos + saved registration) & 5.8\,s & -- \\
Export case (546 IVUS frames, 133\,MB zip) & 1.1\,s & -- \\
\bottomrule\end{tabular}
\end{table}

\begin{figure}[t]\centering
\includegraphics[width=\linewidth]{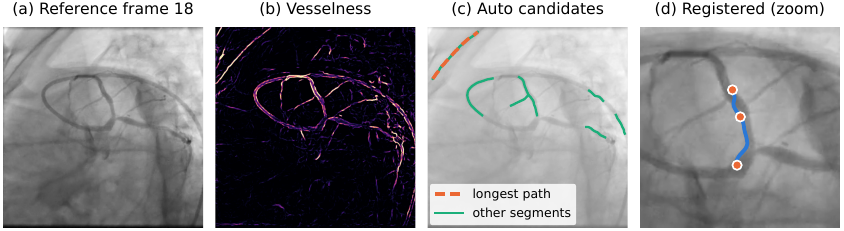}
\caption{Centreline extraction on the clinical case. (a) Reference frame. (b) Frangi vesselness. (c) Automatic candidates: on this frame the default longest-path candidate (dashed orange) follows the guiding catheter, not a coronary artery. (d) The centreline used for registration, covering the pulled-back Left Main segment, with its three anchors (orange).}
\label{fig:pipeline}
\end{figure}

\subsection{Automatic steps help; the clinician decides}
\label{sec:res-phantom}
\begin{figure}[t]\centering
\includegraphics[width=\linewidth]{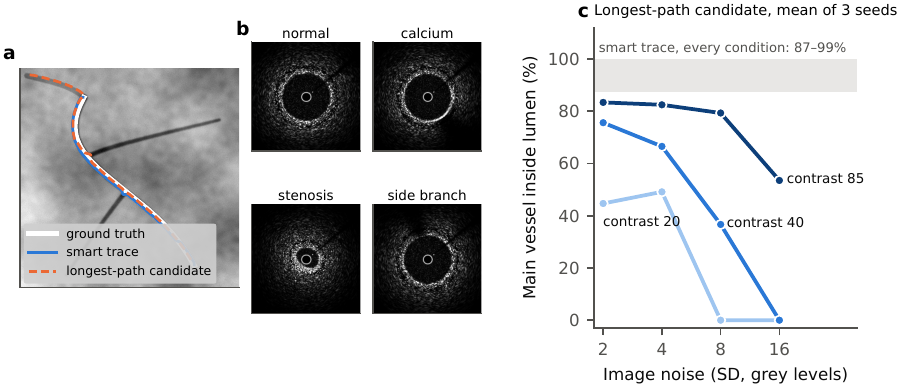}
\caption{Phantom experiment. (a) A phantom reference frame with the true centreline (white), the smart trace (blue) and the automatic longest-path candidate (dashed orange). (b) Frames from the synthetic IVUS pullback used by the public demo. (c) Share of the main vessel covered (inside the lumen) by the automatic longest-path candidate, by image noise and contrast. The grey band shows the smart trace, which stayed at 87--99\% in every condition.}
\label{fig:phantom}
\end{figure}

On phantoms (Table~\ref{tab:phantom}, Figure~\ref{fig:phantom}c), the suggested reference frame was fully opacified in all \PhantomRuns{} phantoms.
The detected opacified window was within 0--2 frames of the truth at the highest contrast. It was within 4 frames at contrast 40 and within 9 frames at contrast 20, where faint early filling is missed.
The smart trace stayed inside the true lumen along \TraceLumenMin{}--\TraceLumenMax{}\% of the vessel (condition means). Only the noisiest, faintest condition fell below 97\%.
Its mean distance from the true centreline was \TraceDistMin{}--\TraceDistMax{} pixels, comparable to the lumen radius (4.5--8 pixels): the minimal path tends to run along one side of a wide, flat vesselness ridge rather than its middle. This offset did not grow with noise and is acceptable for locating IVUS frames along the vessel, but it would bias diameter measurements.

The fully automatic longest-path candidate was much less robust. At the highest contrast it covered \LongestHi{}\% of the main vessel with little noise and \LongestHiNoisy{}\% with heavy noise. At lower contrast it fell to 0\%: the longest path in the skeleton graph then lay outside the main vessel.
Disabling background subtraction barely changed the trace and gave mixed results for the candidates (last column of Table~\ref{tab:phantom}).
The same failure appeared on real data: on the clinical reference frame the longest candidate followed the guiding catheter (Figure~\ref{fig:pipeline}c).

\begin{table}[t]\centering\small
\caption{Automatic steps on angiogram phantoms with known ground truth (mean of 3 noise seeds per row; true lumen radius 4.5--8\,px). \emph{Contrast} is the peak darkening of the opacified vessel in grey levels; the shipped pipeline is ``bgsub''. Window error = largest error of the detected opacified-frame range (frames). Trace distance = mean distance of the smart trace from the true centreline. \emph{In lumen} = share of the true vessel whose nearest traced (or candidate) point lies inside the lumen. The last column repeats the longest-path candidate with background subtraction disabled.}
\label{tab:phantom}
\begin{tabular}{rrrrrrrr}\toprule
& & & \multicolumn{2}{c}{Smart trace} & \multicolumn{3}{c}{Automatic candidates, in lumen (\%)} \\ \cmidrule(lr){4-5}\cmidrule(lr){6-8}
Contrast & Noise SD & Window err. & Dist. (px) & In lumen (\%) & Longest & Best & Longest, no bgsub \\ \midrule
85 & 2 & 2.0 & 4.4 & 99 & 83 & 83 & 77 \\
85 & 4 & 0.0 & 4.5 & 99 & 82 & 82 & 97 \\
85 & 8 & 1.0 & 4.5 & 98 & 79 & 79 & 69 \\
85 & 16 & 0.7 & 4.4 & 98 & 54 & 54 & 33 \\
\addlinespace
40 & 2 & 4.0 & 4.4 & 99 & 76 & 76 & 74 \\
40 & 4 & 3.0 & 4.5 & 98 & 67 & 67 & 68 \\
40 & 8 & 2.0 & 4.3 & 99 & 37 & 37 & 34 \\
40 & 16 & 2.3 & 4.2 & 97 & 0 & 0 & 0 \\
\addlinespace
20 & 2 & 9.0 & 4.3 & 99 & 45 & 48 & 78 \\
20 & 4 & 9.0 & 4.1 & 99 & 49 & 49 & 12 \\
20 & 8 & 7.0 & 3.9 & 99 & 0 & 0 & 2 \\
20 & 16 & 3.3 & 4.6 & 87 & 0 & 0 & 0 \\
\bottomrule\end{tabular}
\end{table}

\subsection{Case study: a registered Left Main pullback}
\label{sec:res-case}
\begin{figure}[t]\centering
\includegraphics[width=\linewidth]{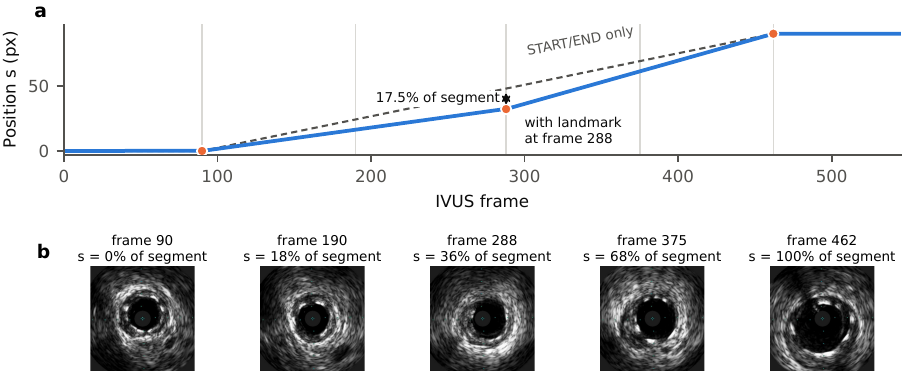}
\caption{Registration of the clinical pullback. (a) Position along the centreline for every IVUS frame: three anchors (orange, solid blue line) against START/END only (dashed). Frames before START and after END keep the end positions. (b) IVUS frames at five positions along the segment, as the export pairs them with the angiogram.}
\label{fig:registration}
\end{figure}

In Case~01 the clinician used angiographic run MF1, frame 18, and pullback MF2 (546 frames).
The pullback moved between frames 90 and 462: \PullbackFrames{} frames, or \PullbackSec{}\,s at 30 frames/s.
The traced segment was \SegLen{} pixels long and ran from the catheter tip along the Left Main, before its bifurcation. In the review sessions it had taken several minutes of discussion to establish that the pullback covered only this short segment.
A third anchor was placed at frame 288.
Compared with a constant-speed assumption between START and END, it moved the position assigned to frame 288 by \RegShift{}\% of the segment (Figure~\ref{fig:registration}). Along the vessel, the probe moved more slowly between frames 90 and 288 than between 288 and 462.
The exported record contains the 546 IVUS frames, each with its position, ready for model training.

\section{Discussion}
\label{sec:discussion}

\subsection{Principal findings}
Registration of archived IVUS pullbacks to the angiogram does not need a vendor console, a server or an installation.
With the clinician choosing the vessel, its direction and its landmarks, an ordinary browser reproduces the original analysis exactly and is fast enough for interactive use.
The automatic steps are useful starting points. The suggested reference frame was always opacified, and the smart trace stayed in the lumen.
They are not a replacement for clinical judgement: the default automatic centreline failed under realistic noise and contrast, and once followed the catheter.
That is why every automatic result in the tool is a proposal.

\subsection{Clinical use}
For the interventional cardiologist the tool answers a practical question on any archived case: \emph{where is this cross-section?}, or the reverse, \emph{what does the wall look like here?}
It supports case review and teaching. Trainees can relate calcium arcs, plaque burden or stent edges to the angiographic silhouette they will see in the catheterisation laboratory, which is useful because angiography often misses calcium that IVUS shows \citep{mintz1995calcification}.
Because nothing is installed or uploaded, a link and a folder of exported videos are enough, and the one-click demo gives a first experience in under a minute.

\subsection{Opportunities for AI}
\label{sec:ai}
Each step of the workflow is a place where a learned model can take over part of the work, and every registered case produces the data to train it (Table~\ref{tab:ai}). Three points cut across all of them.
First, \textbf{the registration is the label}. A registered case pairs every IVUS frame with an angiographic location, the local vessel image and the cardiac phase, which is exactly the supervision that cross-modal models need.
Second, \textbf{inference can stay in the browser}. Models exported to ONNX can run on the user's GPU through ONNX Runtime Web \citep{onnxruntimeweb}, so the privacy property survives the move to deep learning.
Third, \textbf{the clinician stays in the loop}. Model proposals appear as editable candidates, and every correction becomes new training data, the classic setting for active learning \citep{settles2009active}.

\begin{table}[t]\centering\small
\caption{Where AI can enter the workflow. ``Supplies'' is what the tool already records for training or evaluation.}
\label{tab:ai}
\begin{tabularx}{\linewidth}{>{\raggedright\arraybackslash}p{20mm}>{\raggedright\arraybackslash}X>{\raggedright\arraybackslash}X>{\raggedright\arraybackslash}p{34mm}}\toprule
Step & Model & Tool supplies & Expected gain \\ \midrule
Reference frame & Learn end-diastole and opacification from the run, without ECG & Frame chosen by the clinician, contrast and motion curves & Removes the heuristic's odd picks (\S\ref{sec:limits}) \\
Centreline & Coronary segmentation and branch labelling \citep{ronneberger2015unet,yang2019angiosegmentation}, replacing vesselness & Corrected centrelines on real frames; phantoms with exact truth & Robust candidates; no catheter confusion \\
Registration & Detect side branches and stent edges on IVUS and bifurcations on the angiogram, then propose anchors & Anchors with landmark names & Near-automatic registration \\
IVUS tagging & Lumen, vessel and plaque segmentation \citep{bajaj2021ivusseg}; calcium-arc and tissue classification & Frame-range tags; vendor tissue-colour stills in some exports & Automatic plaque profile along the vessel \\
Virtual IVUS & Predict the IVUS cross-section, or its calcium, burden and tissue, from the angiogram at a chosen point \citep{isola2017pix2pix,ho2020ddpm} & Paired (angiogram location, cardiac phase, IVUS frame) for every frame & Plaque information from angiography alone, analogous to angio-derived FFR \citep{tu2016favor} \\
Quality control & Flag implausible maps: speed jumps, non-monotone landmarks, mismatched vessels & Full per-frame map and anchors & Auditable data sets \\
\bottomrule\end{tabularx}
\end{table}

The most ambitious target is \emph{virtual IVUS}: given a contrast-filled angiogram and a point on a vessel, predict what IVUS would show there.
It is not known whether the angiogram contains enough information. Wall calcium is sometimes visible, and contrast dynamics over the cardiac cycle may carry more.
The tool makes the question testable. A model can be conditioned on the angiographic patch around $s$, its position along the vessel and the run's time series, and evaluated against the IVUS frame the registration pairs with $s$.
Simpler targets come first and are clinically useful in their own right: calcium present or absent, calcium arc, and plaque burden.

\subsection{Limitations}
\label{sec:limits}
\begin{itemize}[nosep]
\item \textbf{One clinical case.} Real-data results come from one Left Main case. We have not measured inter-observer agreement, time to register, or accuracy against vendor co-registration or 3D reconstruction \citep{wahle1999fusion,tu2011fusion}. A multi-reader study on a larger set of cases is the next step.
\item \textbf{Simple phantoms.} They are two-dimensional, have no foreshortening or overlap and no realistic contrast physics, and their IVUS is not physically linked to the angiogram. They test the automatic steps, not registration accuracy.
\item \textbf{Single-plane geometry.} Positions are measured along a projected centreline. Foreshortened segments appear shorter, and cardiac and respiratory motion of the vessel relative to the frozen frame is not modelled. Calibration to millimetres is manual.
\item \textbf{Heuristic phase suggestion.} Without ECG it sometimes suggested an end-systole before end-diastole, or a frame at the end of a run (Table~\ref{tab:fidelity}, MF2). The user always confirms the frame.
\item \textbf{Mapping model.} Piecewise-linear interpolation between anchors assumes constant speed within each interval. Non-uniform motion between anchors, including cardiac-cycle oscillation of the probe, is not captured.
\item \textbf{Platform.} The browser edition needs a current Chrome, Edge or Safari, and MP4/MOV input. Frames are held in memory (about 130--220\,MB for 546--895 IVUS frames), and the first visit downloads about 40\,MB of Python packages.
\end{itemize}

\section{Conclusion}
Registering IVUS to the angiogram can be a clinician-led task in an ordinary browser, on any archived case, with no installation and no data leaving the computer.
The browser edition reproduces the validated Python pipeline exactly. Its automatic steps speed the work, but their failures show why the clinician must remain in control.
Every registered case becomes a frame-level angiography--IVUS pair. These pairs are the material needed to find out whether the angiogram alone can tell us what IVUS would see.

\section*{Data and code availability}
The tool is available at \site{}; \textbf{Load demo} opens Case~01 with its registration.
It runs in the user's browser and needs no account or installation. The source code is not publicly released. The measurements behind every table and figure are reported in full in the paper.

\section*{Use of AI assistance}
Software development (including the browser port), experiment orchestration and manuscript drafting were assisted by an AI coding assistant (Claude, Anthropic). The authors designed the study, reviewed all code and results, and take responsibility for the content.

\bibliography{refs}

@article{mintz2001acc,
  author  = {Mintz, Gary S. and Nissen, Steven E. and Anderson, William D. and Bailey, Steven R. and Erbel, Raimund and Fitzgerald, Peter J. and Pinto, Fausto J. and Rosenfield, Kenneth and Siegel, Robert J. and Tuzcu, E. Murat and Yock, Paul G.},
  title   = {American {College} of {Cardiology} Clinical Expert Consensus Document on Standards for Acquisition, Measurement and Reporting of Intravascular Ultrasound Studies ({IVUS})},
  journal = {Journal of the American College of Cardiology},
  volume  = {37}, number = {5}, pages = {1478--1492}, year = {2001}
}

@article{raber2018eapci,
  author  = {R{\"a}ber, Lorenz and Mintz, Gary S. and Koskinas, Konstantinos C. and others},
  title   = {Clinical use of intracoronary imaging. {Part} 1: guidance and optimization of coronary interventions. {An} expert consensus document of the {European} {Association} of {Percutaneous} {Cardiovascular} {Interventions}},
  journal = {European Heart Journal},
  volume  = {39}, number = {35}, pages = {3281--3300}, year = {2018}
}

@article{mintz1995calcification,
  author  = {Mintz, Gary S. and Popma, Jeffrey J. and Pichard, Augusto D. and Kent, Kenneth M. and Satler, Lowell F. and Chuang, Y. Chris and Ditrano, Christopher J. and Leon, Martin B.},
  title   = {Patterns of calcification in coronary artery disease. {A} statistical analysis of intravascular ultrasound and coronary angiography in 1155 lesions},
  journal = {Circulation},
  volume  = {91}, number = {7}, pages = {1959--1965}, year = {1995}
}

@article{zhang2018ultimate,
  author  = {Zhang, Junjie and Gao, Xiaofei and Kan, Jing and others},
  title   = {Intravascular Ultrasound Versus Angiography-Guided Drug-Eluting Stent Implantation: The {ULTIMATE} Trial},
  journal = {Journal of the American College of Cardiology},
  volume  = {72}, number = {24}, pages = {3126--3137}, year = {2018}
}

@article{lee2023renovate,
  author  = {Lee, Joo Myung and Choi, Ki Hong and Song, Young Bin and others},
  title   = {Intravascular Imaging-Guided or Angiography-Guided Complex {PCI}},
  journal = {New England Journal of Medicine},
  volume  = {388}, number = {18}, pages = {1668--1679}, year = {2023}
}

@article{wahle1999fusion,
  author  = {Wahle, Andreas and Prause, Guido P. M. and DeJong, Steven C. and Sonka, Milan},
  title   = {Geometrically correct 3-{D} reconstruction of intravascular ultrasound images by fusion with biplane angiography---methods and validation},
  journal = {IEEE Transactions on Medical Imaging},
  volume  = {18}, number = {8}, pages = {686--699}, year = {1999}
}

@article{tu2011fusion,
  author  = {Tu, Shengxian and Holm, Niels R. and Koning, Gerhard and Huang, Zheng and Reiber, Johan H. C.},
  title   = {Fusion of {3D} {QCA} and {IVUS/OCT}},
  journal = {The International Journal of Cardiovascular Imaging},
  volume  = {27}, number = {2}, pages = {197--207}, year = {2011}
}

@article{tu2016favor,
  author  = {Tu, Shengxian and Westra, Jelmer and Yang, Junqing and others},
  title   = {Diagnostic Accuracy of Fast Computational Approaches to Derive Fractional Flow Reserve From Diagnostic Coronary Angiography: The International Multicenter {FAVOR} Pilot Study},
  journal = {JACC: Cardiovascular Interventions},
  volume  = {9}, number = {19}, pages = {2024--2035}, year = {2016}
}

@inproceedings{frangi1998vesselness,
  author    = {Frangi, Alejandro F. and Niessen, Wiro J. and Vincken, Koen L. and Viergever, Max A.},
  title     = {Multiscale vessel enhancement filtering},
  booktitle = {Medical Image Computing and Computer-Assisted Intervention ({MICCAI}'98)},
  series    = {Lecture Notes in Computer Science}, volume = {1496}, pages = {130--137},
  publisher = {Springer}, year = {1998}
}

@article{zhang1984thinning,
  author  = {Zhang, T. Y. and Suen, Ching Y.},
  title   = {A fast parallel algorithm for thinning digital patterns},
  journal = {Communications of the ACM},
  volume  = {27}, number = {3}, pages = {236--239}, year = {1984}
}

@article{dijkstra1959,
  author  = {Dijkstra, Edsger W.},
  title   = {A note on two problems in connexion with graphs},
  journal = {Numerische Mathematik},
  volume  = {1}, pages = {269--271}, year = {1959}
}

@article{cohen1997minimal,
  author  = {Cohen, Laurent D. and Kimmel, Ron},
  title   = {Global minimum for active contour models: a minimal path approach},
  journal = {International Journal of Computer Vision},
  volume  = {24}, number = {1}, pages = {57--78}, year = {1997}
}

@article{kruskal1956,
  author  = {Kruskal, Joseph B.},
  title   = {On the shortest spanning subtree of a graph and the traveling salesman problem},
  journal = {Proceedings of the American Mathematical Society},
  volume  = {7}, number = {1}, pages = {48--50}, year = {1956}
}

@inproceedings{ronneberger2015unet,
  author    = {Ronneberger, Olaf and Fischer, Philipp and Brox, Thomas},
  title     = {U-Net: Convolutional Networks for Biomedical Image Segmentation},
  booktitle = {Medical Image Computing and Computer-Assisted Intervention ({MICCAI} 2015)},
  series    = {Lecture Notes in Computer Science}, volume = {9351}, pages = {234--241},
  publisher = {Springer}, year = {2015}
}

@article{yang2019angiosegmentation,
  author  = {Yang, Su and Kweon, Jihoon and Roh, Jae-Hyung and others},
  title   = {Deep learning segmentation of major vessels in X-ray coronary angiography},
  journal = {Scientific Reports},
  volume  = {9}, pages = {16897}, year = {2019}
}

@article{bajaj2021ivusseg,
  author  = {Bajaj, Retesh and Huang, Xingru and Kilic, Yakup and others},
  title   = {Advanced deep learning methodology for accurate, real-time segmentation of high-resolution intravascular ultrasound images},
  journal = {International Journal of Cardiology},
  volume  = {339}, pages = {185--191}, year = {2021}
}

@inproceedings{isola2017pix2pix,
  author    = {Isola, Phillip and Zhu, Jun-Yan and Zhou, Tinghui and Efros, Alexei A.},
  title     = {Image-to-Image Translation with Conditional Adversarial Networks},
  booktitle = {Proceedings of the IEEE Conference on Computer Vision and Pattern Recognition ({CVPR})},
  pages     = {1125--1134}, year = {2017}
}

@inproceedings{ho2020ddpm,
  author    = {Ho, Jonathan and Jain, Ajay and Abbeel, Pieter},
  title     = {Denoising Diffusion Probabilistic Models},
  booktitle = {Advances in Neural Information Processing Systems},
  volume    = {33}, pages = {6840--6851}, year = {2020}
}

@techreport{settles2009active,
  author      = {Settles, Burr},
  title       = {Active Learning Literature Survey},
  institution = {University of Wisconsin--Madison, Department of Computer Sciences},
  number      = {1648}, year = {2009}
}

@article{fedorov2012slicer,
  author  = {Fedorov, Andriy and Beichel, Reinhard and Kalpathy-Cramer, Jayashree and others},
  title   = {{3D} {Slicer} as an image computing platform for the {Quantitative} {Imaging} {Network}},
  journal = {Magnetic Resonance Imaging},
  volume  = {30}, number = {9}, pages = {1323--1341}, year = {2012}
}

@article{vanderwalt2014skimage,
  author  = {van der Walt, St{\'e}fan and Sch{\"o}nberger, Johannes L. and Nunez-Iglesias, Juan and Boulogne, Fran{\c{c}}ois and Warner, Joshua D. and Yager, Neil and Gouillart, Emmanuelle and Yu, Tony and {the scikit-image contributors}},
  title   = {scikit-image: image processing in {Python}},
  journal = {PeerJ},
  volume  = {2}, pages = {e453}, year = {2014}
}

@article{virtanen2020scipy,
  author  = {Virtanen, Pauli and Gommers, Ralf and Oliphant, Travis E. and others},
  title   = {{SciPy} 1.0: fundamental algorithms for scientific computing in {Python}},
  journal = {Nature Methods},
  volume  = {17}, pages = {261--272}, year = {2020}
}

@article{harris2020numpy,
  author  = {Harris, Charles R. and Millman, K. Jarrod and van der Walt, St{\'e}fan J. and others},
  title   = {Array programming with {NumPy}},
  journal = {Nature},
  volume  = {585}, pages = {357--362}, year = {2020}
}

@article{bradski2000opencv,
  author  = {Bradski, Gary},
  title   = {The {OpenCV} Library},
  journal = {Dr. Dobb's Journal of Software Tools},
  year    = {2000}
}

@misc{pyodide,
  author       = {{The Pyodide development team}},
  title        = {Pyodide: {Python} with the scientific stack, compiled to {WebAssembly}},
  howpublished = {\url{https://pyodide.org}},
  note         = {Version 0.27.7}, year = {2025}
}

@misc{webcodecs,
  author       = {{W3C}},
  title        = {{WebCodecs}},
  howpublished = {W3C Working Draft, \url{https://www.w3.org/TR/webcodecs/}},
  year         = {2025}
}

@misc{mp4boxjs,
  author       = {{GPAC}},
  title        = {mp4box.js: {JavaScript} version of {GPAC}'s {MP4Box} tool},
  howpublished = {\url{https://github.com/gpac/mp4box.js}},
  note         = {Version 0.5.2}, year = {2023}
}

@misc{onnxruntimeweb,
  author       = {{Microsoft}},
  title        = {{ONNX} {Runtime} {Web}},
  howpublished = {\url{https://onnxruntime.ai}},
  year         = {2025}
}

\appendix
\section{Quick-start guide for clinicians}
\label{app:guide}
\begin{enumerate}[nosep]
\item Open \site{} in Chrome or Edge. Wait for \emph{engine ready} (about 30\,s on the first visit only).
\item Press \textbf{Load demo}, then press $\leftarrow$/$\rightarrow$ to step through the IVUS and watch the marker move, or click the blue centreline to jump the IVUS.
\item For your own case, export the runs as MP4, click \textbf{Browse\ldots} $\to$ \textbf{Open folder\ldots}, and pick the angiogram and the pullback.
\item \textbf{Analyze contrast \& cardiac phase}, check the suggested frame, \textbf{Freeze this frame as reference}.
\item \textbf{Auto-detect centerlines} and \textbf{Use selected}, or \textbf{Smart trace} (click start, then end). Trim to the pulled-back segment; flip if $s=0$ is not the pullback start.
\item On the IVUS, go to the first moving frame and press \textbf{Set pullback START}, then to the last and press \textbf{Set pullback END}.
\item Add landmarks: go to a frame showing a side branch, press \textbf{Add anchor} (or A), and drag the new dot on the mapping chart until its dot on the angiogram sits on the branch.
\item Tag segments if wanted, then press \textbf{Export case}. A zip appears in Downloads; \textbf{Load case\ldots} reopens it.
\end{enumerate}

\begin{figure}[h]\centering
\includegraphics[width=.72\linewidth]{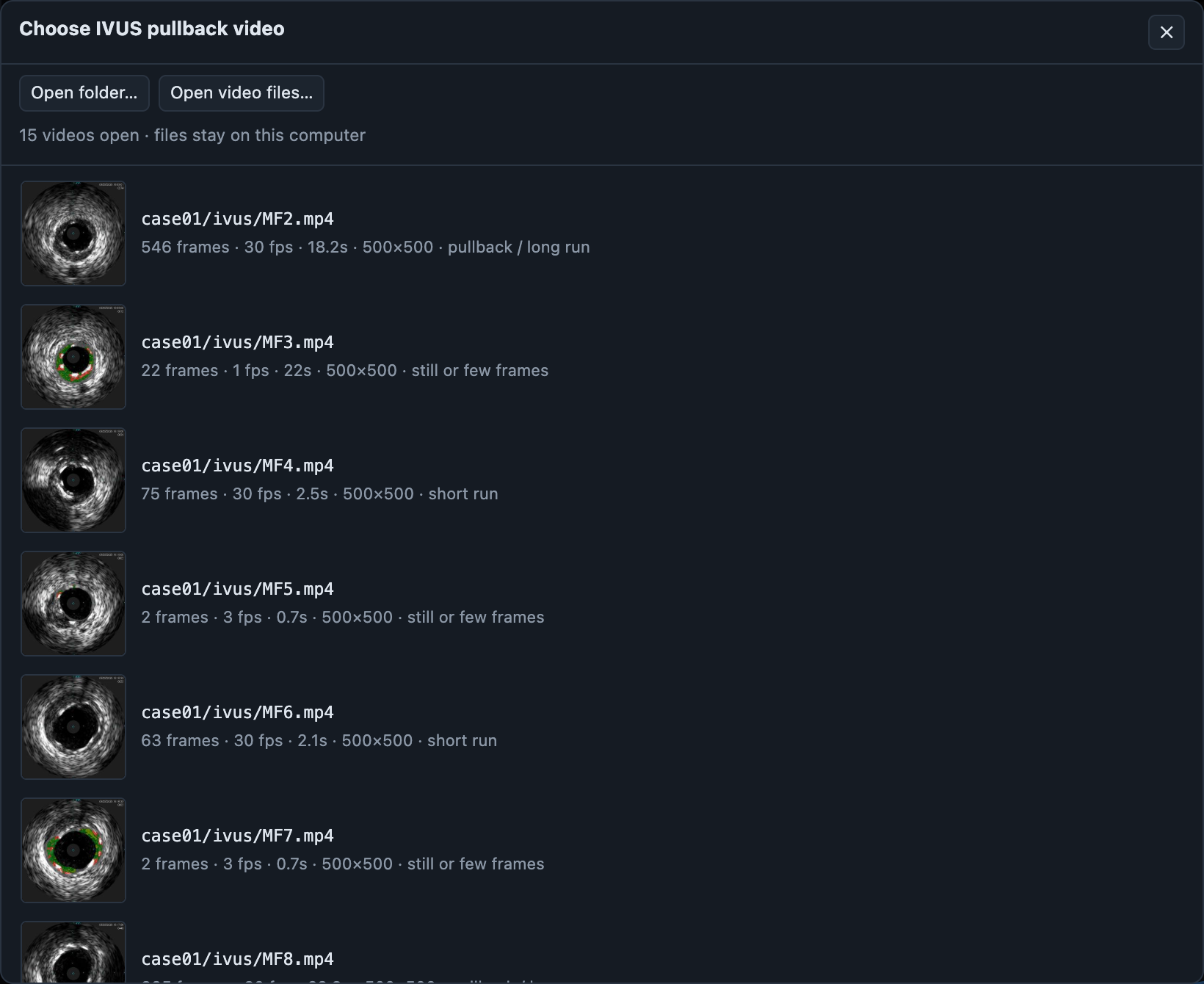}
\caption{Choosing videos. Every series of the case is listed with a thumbnail, frame count, frame rate, duration and a coarse label, so the pullback (here MF2, 546 frames) is easy to find. Some exports are colour-coded tissue stills (MF3, MF7).}
\label{fig:browse}
\end{figure}

\begin{figure}[h]\centering
\includegraphics[width=\linewidth]{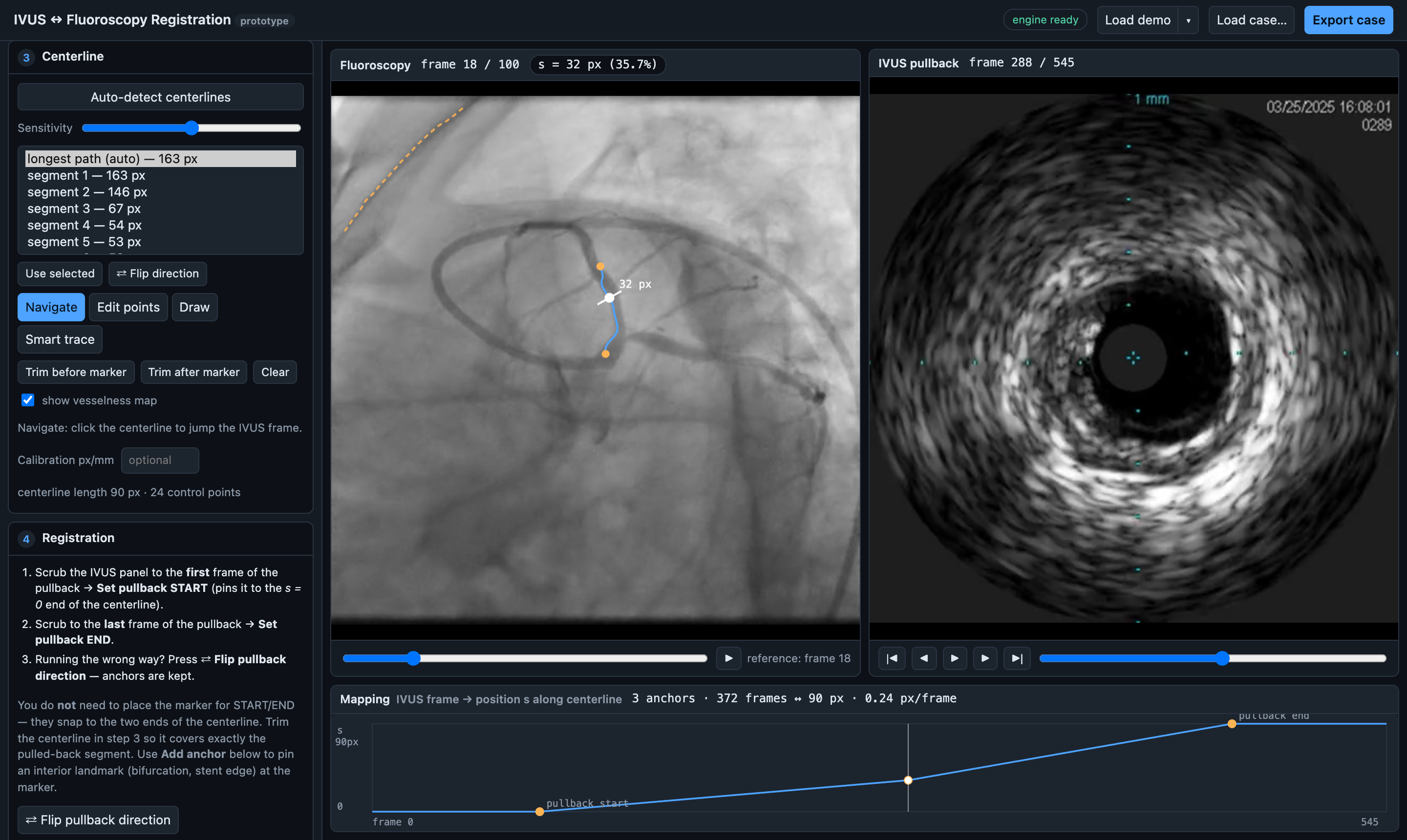}
\caption{Step 3 on the clinical case: the automatic candidates are listed by length, and the default longest path (dashed, top left) follows the guiding catheter. The registered centreline from the saved case is shown in blue.}
\label{fig:ui-centerline}
\end{figure}

\begin{figure}[h]\centering
\includegraphics[width=\linewidth]{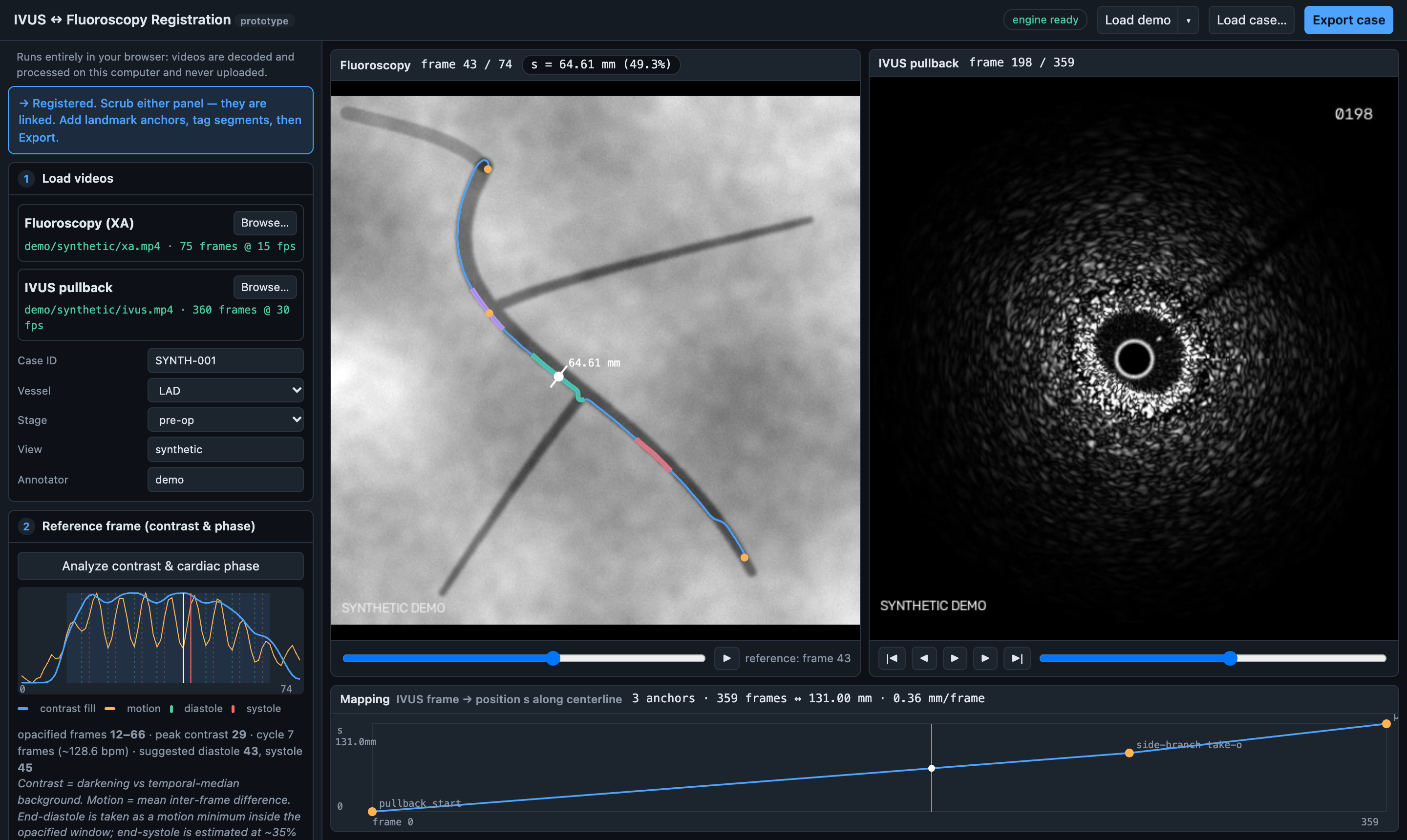}
\caption{The synthetic demonstration case (no patient data). Tagged IVUS segments are drawn on the centreline in colour: calcium (red), fibrous stenosis (teal), bifurcation (violet). The mapping bends at the side-branch landmark. The IVUS shows the stenosis frame.}
\label{fig:synthetic}
\end{figure}

\section{Export manifest}
\label{app:manifest}
Abridged manifest of Case~01 (arrays shortened):
\begin{lstlisting}
{ "case_id": "CASE-01", "vessel": "Left Main", "procedure_stage": "pre-op",
  "fluoro_reference_frame": {"rel": "case01/xa/MF1.mp4", "frame_index": 18,
                             "image": "fluoro_ref.png", "fps": 15, "size": [500, 500]},
  "ivus_video": {"rel": "case01/ivus/MF2.mp4", "n_frames": 546, "fps": 30,
                 "frames_dir": "ivus_frames"},
  "centerline": {"points_xy": [[x, y], ...], "arc_length": [0.0, ...],
                 "total_length": 90.149, "origin": "pullback_start",
                 "arc_length_units": "px", "calibration": null},
  "registration": {
    "anchors": [
      {"ivus_frame": 90,  "s": 0.0,    "s_fraction": 0.0,      "landmark": "pullback start"},
      {"ivus_frame": 288, "s": 32.22,  "s_fraction": 0.357405, "landmark": null},
      {"ivus_frame": 462, "s": 90.151, "s_fraction": 1.000017, "landmark": "pullback end"}],
    "mapping_model": "monotonic_piecewise_linear",
    "frame_to_position": [{"frame": 0, "s": 0.0}, ..., {"frame": 545, "s": 90.151}]},
  "segment_tags": [],
  "annotator": "", "timestamp": "2026-11-17T02:44:46Z" }
\end{lstlisting}

\section{Demo files}
\label{app:demo}
A demo is a JSON file listing the videos (a URL relative to the file, or a path matched in a folder the user opens) and the settings to apply: case metadata, whether to run the analysis, the reference frame, the centreline (points, or an automatic candidate's index), calibration, anchors as (frame, fraction), tags and the starting IVUS frame. \texttt{?demo=<id>} in the address opens a bundled demo; \textbf{Save current as demo\ldots} writes one from the current state.
A research group can therefore publish a teaching case, or a benchmark registration, as a single link.

\end{document}